\documentclass{article} 
\usepackage[preprint]{iclr2026_conference}
\usepackage{times}

\usepackage{amsmath,amsfonts,bm}

\def\eqref#1{equation~\ref{#1}}

\def\1{\bm{1}}

\DeclareMathAlphabet{\mathsfit}{\encodingdefault}{\sfdefault}{m}{sl}
\SetMathAlphabet{\mathsfit}{bold}{\encodingdefault}{\sfdefault}{bx}{n}

\usepackage[utf8]{inputenc}
\usepackage[T1]{fontenc}
\usepackage{amsmath, amssymb, amsfonts, amsthm, mathtools}
\usepackage{algorithm}
\usepackage{algpseudocode}
\usepackage{enumitem}
\usepackage[most]{tcolorbox}
\usepackage{url}
\usepackage{booktabs}
\usepackage{nicefrac}
\usepackage{microtype}
\usepackage{xcolor}
\setcitestyle{numbers,square}
\usepackage{hyperref}
\usepackage{bm}
\usepackage{pgfplots}
\pgfplotsset{compat=1.16}
\usepgfplotslibrary{fillbetween}   

\theoremstyle{plain}
\newtheorem{theorem}{Theorem}[section]

\newtheorem{proposition}[theorem]{Proposition}

\theoremstyle{definition}
\newtheorem{assumption}{Assumption}[section]
\newtheorem{definition}{Definition}[section]

\title{Planning as Emergent Behavior in Reinforcement Learning with Relational Hidden States}

\author{{Armin Sommer} \\
{ETH Zurich, Switzerland}\\
{\texttt{sommera@ethz.ch}}}

\begin{document}

\maketitle

\begin{abstract}
Reinforcement learning is conventionally divided into \emph{model-based} and \emph{model-free} methods. In this taxonomy, model-based methods perform lookahead planning over a learned world model, whereas model-free methods learn a reactive state–action mapping. Recent work, however, has shown that planning can emerge from model-free reinforcement learning alone. The conditions under which this behavior emerges from a pure reward-maximization objective have so far remained unclear. In this paper, we present evidence that, in the observed cases, the hidden-state structure of the neural architecture is the deciding factor. We find that a network of relational hidden states, each anchored to an environment state and exchanging messages along learned relations, acquires a planning mechanism. These hidden states recover the environment’s transition structure in their learned relations, and improve the policy at decision time by planning over the learned graph. In a matched control agent that must additionally discover which cells represent which states, no such binding arises, and no planning follows from it. We argue that this explains the observed phenomenon of emergent planning in model-free reinforcement learning and raises the question of how common such emergent planning might be more generally. Finally, we hypothesize that the discovered mechanism could describe how planning emerges from pure reward maximization in the human brain through a neural architectural prior.
\end{abstract}

\section{Introduction}

Reinforcement learning (RL) methods are conventionally divided into two families
based on whether the agent learns the dynamics of its environment
\citep{sutton2018reinforcement}. \emph{Model-based} methods learn, or are
supplied with, a model of the transition dynamics and use it to \emph{plan}: at
decision time the agent searches over the model, simulating the consequences of
candidate action sequences and selecting an action on that basis. Examples include
tree-search agents such as MuZero \citep{schrittwieser2020mastering} and its
predecessor AlphaGo Zero \citep{silver2017mastering}.
\emph{Model-free} methods instead learn their behavior directly
from rewards without explicitly modelling their environment’s dynamics. The resulting policy is a reactive mapping
from states to actions: it is evaluated in a single forward pass, simulates no
future, and represents no knowledge of the dynamics. Under this taxonomy, planning
is the province of the model-based family, and the model-free agent is
characterized precisely by its absence.

However, it has been observed that model-free agents can nonetheless learn to
plan. \citet{guez2019investigation} train a Deep Repeated ConvLSTM (DRC), a
generic recurrent network composed of convolutional hidden states with no
structure specific to planning. Using a standard model-free RL algorithm, they find
that it exhibits the characteristics typically associated with a model-based
planner: it generalizes across combinatorial, irreversible state spaces, is
data-efficient, and, most diagnostically, improves its performance when granted
additional test-time computation, or ``thinking time,'' before acting. On
combinatorial domains such as Sokoban, it matches or exceeds model-based agents.
These behavioural signatures indicate that the model-based/model-free boundary is
less sharp than the taxonomy implies: a network with no explicit planning machinery can
come to behave as though it deliberates.

Subsequent work has established that this behavior reflects an internal planning
process rather than a collection of reactive heuristics. Using concept-based
probing, \citet{bush2025interpreting} decode from a Sokoban-playing DRC two
spatially-localized, square-level quantities, namely the direction from which the
agent will next enter a square and the direction in which it will next push a
box, and show these assemble into coherent planning trajectories. The authors find that these representations are progressively refined over the network's internal computation steps, that they transfer to out-of-distribution levels, and that intervening on them
causally redirects the agent's behavior. Finally, they show that the decoded procedure resembles a
parallelized bidirectional search rather than the forward rollouts of
conventional model-based planners. \citet{taufeeque2024planning} reach a
consistent conclusion by reverse-engineering a similar convolutional recurrent
agent, identifying an iterative computation that propagates plan information
across the spatial state, later refined into a description in terms of path
channels and plan-extension kernels \citep{taufeeque2025pathchannels}. Moreover, the
phenomenon is not confined to recurrent
architectures: \citet{bush2025interpreting} report the same planning signatures
in a feedforward $24$-layer ResNet, whose plan representations sharpen across
successive layers as the DRC's sharpen across recurrent iterations, consistent
with the view that residual networks perform unrolled iterative estimation
\citep{greff2017highway}.

Taken together, these results demonstrate that planning can emerge under
model-free reinforcement learning and characterize \emph{what} algorithm a particular trained
agent comes to implement. They leave open a more general question:
under what conditions, and in particular owing to which property of the function
approximator, does planning emerge at all? \citet{bush2025interpreting} raise
precisely this question and leave it unresolved. The present work addresses it.

We identify the responsible property as a \emph{relational hidden state}: a
recurrent state organized as a set of cells whose updates depend on learned
pairwise relations among them. Convolution and attention are two instances of such a relational, graph-structured operation: convolution relates each cell to a fixed local neighbourhood, whereas attention relates cells through learned, content-dependent weights; both are expressible as message passing over a graph defined on the cells \citet{battaglia2018relational}. Relational structure of this kind has previously been used to improve reasoning
in reinforcement learning \citep{zambaldi2018relational} and, more broadly, to
learn the interactions and dynamics of structured systems
\citep{sanchez2018graph,kipf2018neural}.

We hypothesize that emergent planning requires one architectural prior: an \emph{anchoring} that binds each environment state to its own hidden cell. Given such binding, the messages exchanged between anchored cells can come to respect the environment's transition relation, and successive updates then propagate decision-relevant information along the induced state graph: one hop per update in the convolutional case, a global refinement of the whole field in the attention case. Either way, the aggregate computation is a multi-step lookahead in latent space that deepens with each update, accounting for the gains these agents show under added test-time computation.

In this work we test this hypothesis directly. We train an attention-based agent whose relational core is a dense, unmasked attention recurrence, no
cell is privileged as another's neighbour, so that the transition graph, rather
than being wired in by a convolution, must be discovered from reward alone. Using
mechanistic probes and causal interventions we show that this agent binds board
squares to cells, recovers the transition graph in its learned attention, lays out
a goal-directed plan over that graph, and refines it with each thinking step. We
then isolate the responsible ingredient with a matched control that
must additionally learn its binding: no stable binding forms, no transition graph
settles into the relations, and planning does not emerge.

Read together, the established convolutional results and the new attention
results we present show that the same planning computation arises whether the
transition graph is given or learned. A third agent (\S\ref{sec:emp-slots})
completes the argument: when the given binding is removed, no binding forms,
no transition graph settles into the relations, and no planning emerges. It is
therefore the relational bias \emph{operating over state-bound cells}, not any
built-in graph, that produces planning.

\section{Background}
\label{sec:prelim}

\subsection{Reinforcement Learning}

We consider a deterministic Markov decision process (MDP)
$(\mathcal S,\mathcal A,\mathcal T,\mathcal R)$
\citep{puterman1994markov} with a finite state set
$\mathcal S$, a finite action set $\mathcal A$, a deterministic transition map
$\mathcal T:\mathcal S\times\mathcal A\to\mathcal S$, and a reward function
$\mathcal R:\mathcal S\to\mathbb R$. The objective of reinforcement learning is to find a policy $\pi$ that maximizes the expected discounted return $\mathbb E_\pi\bigl[\sum_{i\ge 0}\gamma^{i}\,r_{t+i}\mid s_t=s\bigr]$,
with discount factor $\gamma\in(0,1)$. \emph{Model-free} methods optimize this
objective directly, without learning the transition map. Most modern methods use
an actor--critic scheme \citep{konda1999actor}: a critic
\begin{align} V^\pi(s)=\mathbb E_\pi\bigl[\sum_{i\ge 0}\gamma^{i}\,r_{t+i}\mid s_t=s\bigr] \end{align}
estimates the return and is learned by temporal-difference (TD) learning
\citep{sutton1988learning}, while the parameters of the actor $\pi_\phi$ are
updated by the policy gradient \citep{sutton2000policy}
\begin{align}
  \nabla_\phi J(\phi) = \mathbb E_\pi\Bigl[\sum_{t}
  \nabla_\phi\log\pi_\phi(a_t\mid s_t)\,\delta_t\Bigr],
\end{align}
where $\delta_t$ is an advantage estimator, such as the $n$-step TD error or the
generalized advantage estimate \citep{schulman2015high}. Scalable
implementations of this scheme, such as
IMPALA \citep{espeholt2018impala}, are the standard training substrate for the
architectures of this paper.

\subsection{Decision-time Planning}
\label{sec:dtp}
In this paper, \emph{planning} refers to \emph{decision-time planning}: selecting
the action at the current state by simulating and evaluating the consequences of
future actions \citep{sutton2018reinforcement}. In the model-based setting these
consequences are simulated under a learned model
$\mathcal{M}=(\hat{p},\hat{r})$, where $\hat{p}(s'\mid s,a)$ approximates the
transition dynamics and $\hat{r}(s,a)$ the reward
\citep{ha2018world,hafner2019planet,hafner2023dreamerv3}. Such models are
typically fit against a predictive loss, which need not align with control
performance, creating the objective-mismatch problem \citep{lambert2020objective}.

Rather than storing an explicit policy, decision-time planning defines the action
at the current state $s_t$ \emph{implicitly}, by optimizing a simulated lookahead
of horizon $K$ at the moment the action is required:
\begin{align}
  \pi(s_t) \;=\; \operatorname*{arg\,max}_{a_t}\;
  \mathbb{E}_{s_{t+k+1} \sim \hat{p}(\cdot\mid s_{t+k},a_{t+k})}\!\left[
    \sum_{k=0}^{K-1}\gamma^{k}\,\hat{r}(s_{t+k},a_{t+k})
    \;+\;\gamma^{K}\,V^\pi(s_{t+K})
    \;\middle|\; s_t, a_t
  \right],
  \label{eq:planning}
\end{align}
where the inner maximization is over the action sequence $a_{t:t+K-1}$ and the
terminal estimate $V^\pi$ bootstraps the return beyond the horizon. Only the
first action $a_t$ is executed; the agent then re-plans from $s_{t+1}$ in
receding-horizon fashion. At $K{=}1$ this reduces to the one-step greedy policy with respect to $V^\pi$; larger $K$ evaluates progressively longer futures under the model.

\section{Relational Hidden States}
\label{sec:core}

\subsection{Neural Architecture}

We study neural networks that move information over their own neural
structure. We view such networks as pools of hidden cells exchanging messages,
with \emph{no} a priori correspondence between
cells and environment states. We refer to this substrate, weight-tied
cells coupled by learned relations, as \emph{relational latent
states}: the inductive bias whose consequences this paper traces
\citep{battaglia2018relational}.

\begin{definition}
[Relational hidden states]
\label{ass:core}
We say that a neural network has relational hidden states if its hidden cells $h_t(i)\in\mathbb R^{n}$ are updated by the other latent cells, indexed by $j$ via
\begin{equation}
  h_t^{k+1}(i)
  \;=\;
  \bigoplus_{j}\,
  F_{\phi}\bigl(h_t^{k}(i),\,h_t^{k}(j)\bigr),
  \label{eq:update}
\end{equation}
where $k \in \{0,...,K-1\}$ is a thinking step and $t$ the environment time. Here, $F_\phi$ is the learned message for the pair $(j\to i)$ and $\bigoplus$ is
any permutation-invariant aggregation over the related cells. The carry
between environment steps is the last thinking state,
$h_t^{0}\equiv h_{t-1}^{K}$.
\end{definition}

Convolutions and the attention mechanism are two instantiations of $F_\phi$.
A \emph{convolution} places the cells on a grid and lets the message depend
only on the sender and its offset to the receiver, with a learned weight
$W({j-i})$ for offset on the kernel support and $0$ otherwise, applied to the content $h_t^{k}(j)$. Those weighted messages are then summed over a local neighborhood. \emph{Attention}
instead lets the message depend on content: a learned score $a_\phi$ between
the two cells weights each sender's latent content $h_t^{k}(j)$ by a linear transformation $W$. Here, the weighted messages are pooled as a
normalized average over all pairs. Content-addressed routing of this kind can be
read as an associative memory \citep{ramsauer2021hopfield} and, when linearized,
as a form of recurrence \citep{katharopoulos2020transformers}, placing
convolution, attention, and external-memory architectures
\citep{graves2016hybrid} on a common relational footing.

\subsection{Hidden-State Binding}

\begin{figure}[t]
\centering
\includegraphics[height=0.3\linewidth]{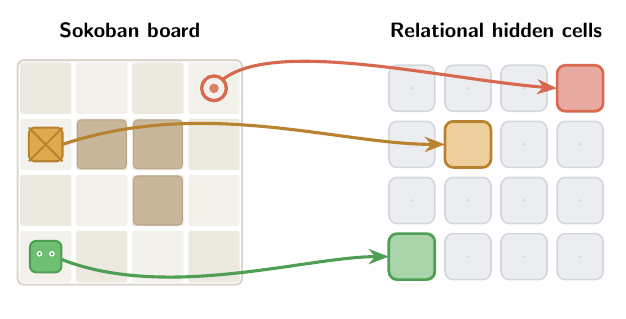}
\caption{\textbf{State-graph binding.} Each Sokoban board square (left) is bound to one relational
hidden cell (right) of the neural network via a fixed indexing (color-matched arrows and cells).}
\label{fig:binding}
\end{figure}

A relational core (Def.~\ref{ass:core}) carries content over a fixed pool of
cells, but nothing in Eq.~\eqref{eq:update} forces those cells to \emph{mean}
anything. Suppose the network computes some statistic $g(s)$ of a state, which we call a
feature of the state. We use \emph{binding} to denote the process by which training turns the cell pool into an internal, content-addressed copy of the states the agent must reason
about, with cells related to one another along the edges the environment permits
(Fig.~\ref{fig:binding}).
This mirrors the notion of relational memory in cognitive neuroscience, in which
distinct experiences are represented as elements linked by their relations rather
than as isolated items \citep{cohen1993memory,eichenbaum2017integration}.

\begin{definition}[State-graph binding]
\label{def:binding}
We say a neural network's core exhibits \emph{state-graph binding} if there
exists a map $\sigma:\mathcal S \to \mathcal C $, from the environment's states to the net's hidden cells, such
that each state's statistic is carried by a unique cell and recoverable from $h_t(\sigma(s))$. Furthermore, the core's
relations \emph{respect the transition graph} if the message in~\eqref{eq:update} from cell $\sigma(s')$ to cell $\sigma(s)$ contributes
only when $s' \in \mathcal N(s) := \{\mathcal T(s,a) : a \in \mathcal A\}$.
\end{definition}

In a convolution, the cell-to-state binding is learned but the transition-respecting
relations are supplied by the fixed local kernel; in attention, both must
be learned from reward alone. With both conditions in place, each thinking step
moves $g$ one hop along $\mathcal N$: after $K$ steps the agent's cell
$\sigma(s_t)$ has gathered statistics from its $K$-hop neighborhood, and the
carry $h_t^{0}\!\equiv\!h_{t-1}^{K}$ hands that information on to the next
environment step, a horizon that recedes step by step and forms the substrate on
which planning can emerge.


\section{Emergent Planning}
\label{sec:planning}

We now argue that these two structural properties, state-graph
binding and transition-respecting relations, can turn the thinking recursion into decision-time planning. The argument proceeds in three
steps. First, we assume that the action statistics localize to the hidden cells. Second, we argue that the learned
relations \emph{are} a transition model, so that no separate model needs to be fit. Third, we show that this computation can represent a horizon-$K$ neural planner of~\eqref{eq:planning} and stochastic gradient descent can discover decision-time planning from a reward-maximization objective.

\subsection{Localized Decision Statistics}
\label{sec:localized}
We first argue that actions are decodable from the relational core.

\begin{assumption}[Localized readout]
\label{ass:readout}
The neural network binds the actions of a state $s_t$ to its corresponding hidden cell, so that after $K$ thinking steps:
\begin{align}
  \pi(\cdot\mid s_t) \;\approx\; \rho_\pi\!\bigl(h_t^{K}(\sigma(s_t))\bigr),
  \label{eq:readout}
\end{align}
where $\rho_\pi$ is a learned readout implemented by the next hidden layer or the actor head.
\end{assumption}

Under state indexing with transition-respecting relations, the cell update~\eqref{eq:update} moves information only along the transition model: a cell's
content aggregates the statistics of states adjacent to the cell's underlying state,
and, because the weights vanish across absent transitions, never draws on states unreachable from it. How far along the graph a single step carries information depends on the instantiation. A convolution
fetches information exactly from the one-hop successors in our grid-construction. For a learned attention
core we find a \emph{soft} reach over successors: per-step
influence decays geometrically in graph distance rather than truncating at
one hop, and is exactly zero only where transitions are
absent. Additional thinking steps
refine the resulting computation globally rather than advancing a
distance-graded front, and it is this refinement that improves the decision
with test-time computation.

\subsection{Implicit Transition Model}
\label{sec:implicit}

Decision-time planning requires a transition model
$\hat p$. We now show that the trained relational core implicitly carries such a transition model, in two
complementary senses.

\textbf{Actions as successors.} In a deterministic MDP, each action available at
$s$ selects a unique successor, so the map
$a \mapsto \mathcal T(s,a)$ is onto $\mathcal N(s)$. A policy over actions is
therefore equivalent to a distribution over successor states,
\begin{align}
  P^\pi(s'\mid s) \;=\; \sum_{a\,:\,\mathcal T(s,a)=s'} \pi(a\mid s),
  \label{eq:successor-policy}
\end{align}
and, conversely, any distribution supported on $\mathcal N(s)$ induces a policy.
Selecting an action is thus equivalent to selecting a neighbour on the transition graph.

\textbf{Relations as routing.} Let $w(j\to i)\ge 0$ denote the weight that the
cell update in~\eqref{eq:update} places on the message from cell $j$ to cell $i$;
for a convolution this is a kernel weight, for attention an attention score. Define
the induced model by
\begin{align}
   \hat p(s'\mid s) \;\propto\; w\bigl(\sigma(s')\to\sigma(s)\bigr).
   \label{eq:implicit-model}
\end{align}
By Definition~\ref{def:binding}, if the relations are transition-respecting then
$w$ places all of its mass on genuine successors, so
$\operatorname{supp}\hat p(\cdot\mid s)\subseteq\mathcal N(s)$. Thus $\hat p$ is a
transition model of the environment read directly off the routing, without being
fit against an explicit prediction loss.

\subsection{Planning as Message Passing}
\label{sec:mp-planning}

We write $F^{K}_{\phi,\omega}$ for the $K$-fold composition of the cell
update~\eqref{eq:update}, with $\omega(j\to i)$ the aggregation weights of
$\bigoplus$ (Eq.~\eqref{eq:implicit-model}), so that
$h_t^{K}=F_{\phi,\omega}^{K}(h_t^{0})$. If each cell
starts from the local statistic $g(s)$ alone, information about distant goals
and obstacles can reach the readout cell only through message passing along
the learned relations: return is earned exactly to the extent that the
relations transport decision-relevant information.

\begin{proposition}[Message passing can implement decision-time planning]
\label{prop:plan}
Under Assumption~\ref{ass:readout} and Definition~\ref{def:binding}, every
policy the architecture expresses is a relational readout of the thinking
recursion routed by $\omega$,
\begin{align}
  \pi(s_t)\;=\;\rho_\pi\!\Bigl(F_{\phi,\omega}^{K}\bigl(\{\,x(g(s')):s'\in\mathcal S\,\}\bigr)\Bigr),
  \label{eq:implicit}
\end{align}
in which each application of $F_{\phi,\omega}$ lets a cell $\sigma(s)$ pull
messages only from the successor cells $\sigma(\mathcal N(s))$ that $\omega$
admits. This computation can implement the horizon-$K$ decision-time planner
of Eq.~\eqref{eq:planning},
\begin{align}
 \pi(s_t) \;=\; \operatorname*{arg\,max}_{a_t}\;
  \mathbb{E}_{s_{t+k+1} \sim \hat{p}(\cdot\mid s_{t+k},a_{t+k})}\!\left[
    \sum_{k=0}^{K-1}\gamma^{k}\,\hat{r}(s_{t+k},a_{t+k})
    \;+\;\gamma^{K}\,V^\pi(s_{t+K})
    \;\middle|\; s_t, a_t
  \right],
  \label{eq:planner-det}
\end{align}
with $\hat p$ given by $\omega$ as in ~\eqref{eq:implicit-model}. There exist parameters
$(\phi,\rho_\pi)$ under which each application of $F_{\phi,\omega}$ transports information, so that after $K$
steps information about the best leaf state at the horizon has propagated along $\omega$ to $\sigma(s_t)$, and
$\rho_\pi$ returns the action toward the successor carrying it.
\end{proposition}

We thus argue that decision-time planning is a solution that gradient descent can discover
from reward alone: the model is amortized into the routing, the lookahead into
the thinking recursion, and neither is supplied in advance.

\section{Empirical Validation}
\label{sec:emp}

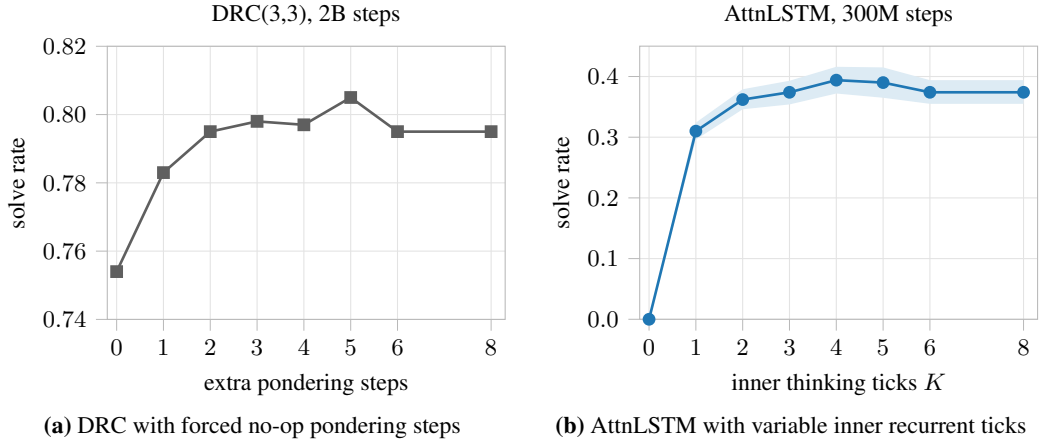
\begin{figure}[t]
\centering
\pgfplotsset{
  ponderaxis/.style={
    width=\linewidth,
    height=5.2cm,
    xmin=-0.2, xmax=8.3,
    xtick={0,1,2,3,4,5,6,8},
    tick align=outside,
    tick pos=left,
    grid=major,
    grid style={gray!22, line width=0.4pt},
    axis line style={gray!55},
    tick style={gray!55},
    tick label style={font=\footnotesize},
    label style={font=\small},
    title style={font=\small, yshift=-2pt},
    every axis plot/.append style={line width=1pt},
  }
}
\definecolor{drcgray}{RGB}{95,95,95}
\definecolor{attnblue}{RGB}{31,119,180}
\begin{minipage}[t]{0.49\textwidth}
\centering
\begin{tikzpicture}
\begin{axis}[
  ponderaxis,
  xlabel={extra pondering steps},
  ylabel={solve rate},
  ymin=0.74, ymax=0.82,
  ytick={0.74,0.76,0.78,0.80,0.82},
  yticklabel style={/pgf/number format/.cd, fixed, fixed zerofill, precision=2},
  title={DRC(3,3), 2B steps},
]
\addplot[mark=square*, drcgray, mark size=1.9pt]
coordinates {
  (0,0.754) (1,0.783) (2,0.795) (3,0.798)
  (4,0.797) (5,0.805) (6,0.795) (8,0.795)
};
\end{axis}
\end{tikzpicture}
\vspace{4pt}
{\footnotesize\textbf{(a)} DRC with forced no-op pondering steps}
\end{minipage}
\hfill
\begin{minipage}[t]{0.49\textwidth}
\centering
\begin{tikzpicture}
\begin{axis}[
  ponderaxis,
  xlabel={inner thinking ticks $K$},
  ylabel={solve rate},
  ymin=0, ymax=0.45,
  ytick={0,0.1,0.2,0.3,0.4},
  yticklabel style={/pgf/number format/.cd, fixed, fixed zerofill, precision=1},
  title={AttnLSTM, 300M steps},
]
\addplot[draw=none,name path=a2up,forget plot] coordinates {(0,0.000)(1,0.324)(2,0.379)(3,0.393)(4,0.416)(5,0.415)(6,0.394)(8,0.394)};
\addplot[draw=none,name path=a2lo,forget plot] coordinates {(0,0.000)(1,0.296)(2,0.346)(3,0.354)(4,0.372)(5,0.365)(6,0.355)(8,0.355)};
\addplot[attnblue!15,forget plot] fill between[of=a2up and a2lo];
\addplot[mark=*, attnblue, mark size=1.9pt]
coordinates {
  (0,0) (1,0.310) (2,0.362) (3,0.374)
  (4,0.394) (5,0.390) (6,0.374) (8,0.374)
};
\end{axis}
\end{tikzpicture}
\vspace{4pt}
{\footnotesize\textbf{(b)} AttnLSTM with variable inner recurrent ticks}
\end{minipage}
\caption{\textbf{Additional test-time computation improves solving.}
Solve rate on held-out Boxoban \texttt{valid\_medium}. \textbf{(a)} For the pretrained
DRC(3,3) of \citet{taufeeque2024planning}, additional computation is provided by
forced no-op pondering steps, following \citet{guez2019investigation}. \textbf{(b)}
For our AttnLSTM, additional computation is provided by increasing the number of
inner recurrent thinking ticks before acting. Both agents improve with extra
test-time recurrent computation, but the horizontal axes represent different
interventions and are therefore shown separately.}
\label{fig:solve}
\end{figure}

The hypothesis advanced so far is that a \emph{relational hidden state}, a pool of cells coupled by learned relations, is the
inductive bias under which decision-time planning emerges from reward alone, and that it does so \emph{without
the environment's transition structure being supplied to the agent}. Existing evidence on convolutional agents
establishes that such planning occurs but cannot isolate this claim: the Deep Repeated ConvLSTM plans on
Sokoban \citep{guez2019investigation}, and its plan has been decoded as square-level, causally-effective,
iteratively-refined concepts \citep{bush2025interpreting, taufeeque2024planning}, yet because a convolution
wires each cell to its spatial neighbours, the transition graph is \emph{built into} the architecture, and the
relational bias there cannot be separated from a hand-given graph. This paper removes that crutch: we present an agent whose relational core is a
\emph{dense, unmasked attention} recurrence, in which no cell is privileged as
another's neighbour and every relation is learned, and we show that decision-time
planning emerges, with the transition graph now discovered from reward.
Whether the relational bias alone suffices, or whether it must operate over a
supplied binding, is exactly what the free-slot control of \S\ref{sec:emp-slots}
isolates.

\paragraph{The two relational cores.}
Both cores instantiate the relational recurrence of definition ~\ref{ass:core}. They differ only in their relational updates. The \emph{convolutional} core is the
Deep Repeated ConvLSTM of \citet{guez2019investigation}: $D{=}3$ ConvLSTM layers whose cells sit on the
$10\times10$ board and exchange messages only with their spatial neighbours through a fixed convolutional
kernel (plus a pooled global channel). In this work, we analyse the
pretrained agent studied by \citet{bush2025interpreting,taufeeque2024planning}. The \emph{attention} core, which we call AttnLSTM, replaces the fixed kernel with
learned, content-based attention: a convolutional encoder maps the board to
$100$ cells (one per square), which each thinking step relates by dense,
multi-head entmax attention with no locality mask. Thus, every hidden cell may attend to every other and which
relations matter must be learned by the agent. These relations are followed by an LSTM gate; $D{=}3$ stacked cells run up to $K{=}6$ thinking steps, the
thinking depth sampled per rollout so the agent must act well at every thinking budget. The AttnLSTM is trained from reward alone (IMPALA
on Boxoban, $300$M steps). Full architectural descriptions and hyperparameters are in
App.~\ref{app:arch}.

\begin{figure}[t]\centering
\begin{minipage}[t]{0.40\textwidth}\centering
\parbox[c][116pt][c]{\linewidth}{\centering
\begin{tikzpicture}[scale=0.6,every node/.style={font=\scriptsize}]
\fill[blue!42](0,4)rectangle(1,5);\fill[blue!28](1,4)rectangle(2,5);\fill[blue!20](2,4)rectangle(3,5);\fill[blue!14](3,4)rectangle(4,5);\fill[blue!10](4,4)rectangle(5,5);
\fill[blue!65](0,3)rectangle(1,4);\fill[blue!42](1,3)rectangle(2,4);\fill[black](2,3)rectangle(3,4);\fill[blue!10](3,3)rectangle(4,4);\fill[blue!7](4,3)rectangle(5,4);
\fill[blue!65](1,2)rectangle(2,3);\fill[black](2,2)rectangle(3,3);\fill[blue!7](3,2)rectangle(4,3);\fill[blue!5](4,2)rectangle(5,3);
\fill[blue!65](0,1)rectangle(1,2);\fill[blue!42](1,1)rectangle(2,2);\fill[black](2,1)rectangle(3,2);\fill[blue!5](3,1)rectangle(4,2);\fill[blue!4](4,1)rectangle(5,2);
\fill[blue!42](0,0)rectangle(1,1);\fill[blue!28](1,0)rectangle(2,1);\fill[black](2,0)rectangle(3,1);\fill[blue!4](3,0)rectangle(4,1);\fill[blue!4](4,0)rectangle(5,1);
\fill[white](0,2)rectangle(1,3);\draw[orange,very thick](0,2)rectangle(1,3);\node at(0.5,2.5){Q};
\draw[gray!40,step=1](0,0)grid(5,5);
\node[white]at(1.5,2.5){\bfseries.24};\node at(3.5,2.5){$\approx$0};\node at(4.5,2.5){$\star$};
\end{tikzpicture}}\\[3pt]
{\small (a) successors a query connects to}
\end{minipage}\hfill
\begin{minipage}[t]{0.58\textwidth}\centering
\parbox[c][116pt][c]{\linewidth}{\centering
\begin{tikzpicture}
\begin{axis}[width=8cm,height=4cm,xlabel={thinking step $k$},
  ylabel={$\|\Delta h\|/\|h\|$ at agent},ylabel style={font=\small},
  xmin=0.6,xmax=12.4,ymin=0,ymax=0.75,xtick={2,4,6,8,10,12},
  legend style={at={(0.5,1.03)},anchor=south,legend columns=-1,draw=none,font=\scriptsize}]
\addplot[draw=none,name path=wonup,forget plot] coordinates {(1,0.235)(2,0.414)(3,0.526)(4,0.606)(5,0.627)(6,0.662)(7,0.673)(8,0.679)(9,0.690)(10,0.692)(11,0.701)(12,0.711)};
\addplot[draw=none,name path=wonlo,forget plot] coordinates {(1,0.136)(2,0.278)(3,0.381)(4,0.452)(5,0.504)(6,0.512)(7,0.536)(8,0.538)(9,0.550)(10,0.556)(11,0.564)(12,0.565)};
\addplot[blue!12,forget plot] fill between[of=wonup and wonlo];
\addplot[draw=none,name path=woffup,forget plot] coordinates {(1,0.105)(2,0.171)(3,0.196)(4,0.222)(5,0.235)(6,0.254)(7,0.263)(8,0.281)(9,0.292)(10,0.298)(11,0.298)(12,0.305)};
\addplot[draw=none,name path=wofflo,forget plot] coordinates {(1,0.066)(2,0.107)(3,0.149)(4,0.181)(5,0.198)(6,0.216)(7,0.228)(8,0.240)(9,0.241)(10,0.245)(11,0.245)(12,0.240)};
\addplot[gray!18,forget plot] fill between[of=woffup and wofflo];
\addplot[mark=*,blue,thick,mark size=1.2pt]
  coordinates {(1,0.186)(2,0.346)(3,0.454)(4,0.529)(5,0.566)(6,0.587)(7,0.604)(8,0.609)(9,0.620)(10,0.624)(11,0.632)(12,0.638)};
\addlegendentry{on-route wall}
\addplot[mark=square*,gray,thick,mark size=1.2pt]
  coordinates {(1,0.085)(2,0.139)(3,0.173)(4,0.202)(5,0.217)(6,0.235)(7,0.245)(8,0.260)(9,0.267)(10,0.271)(11,0.272)(12,0.272)};
\addlegendentry{off-route wall}
\draw[densely dashed,black!45](axis cs:6,0)--(axis cs:6,0.75);
\node[anchor=south,black!55,font=\tiny]at(axis cs:6,0.02){trained depth};
\end{axis}\end{tikzpicture}}\\[3pt]
{\small (b) a route-blocking wall re-plans the move}
\end{minipage}
\caption{\textbf{The attention routing is an amortized transition model.}
\textbf{(a)} A cell $Q$'s settled attention concentrates on feasible one-step
successors ($\approx 0.24$ at one hop) and decays geometrically with
\emph{graph} distance ($\rho\approx 0.61$). Its attention to cells made
unreachable by a wall is approximately zero.
\textbf{(b)} Turning a floor tile on the agent’s route into a wall shifts the current position's hidden cell $2.3\times$ more than turning an off-route
tile at a matched distance into a wall. Additionally, the intervention flips
the agent's greedy move on $19\%$ of route-blocked boards, compared with
$7.2\%$ of off-route boards: a change to the dynamics propagates along the
graph and rewrites the plan.}
\label{fig:wall}
\end{figure}

Previous work established that the DRC improves its solve rate with
additional thinking steps \citep{guez2019investigation}. We find that the attention core shows the same improvement (Fig.~\ref{fig:solve}). However, improvement with computation is
only a necessary signature of planning, not a sufficient one, so it could reflect reactive heuristics sharpened
over iterations as well. We make use of mechanistic interpretability techniques to show that the attention core also implements decision-time
planning over the transition graph. We establish four core properties predicted by the theoretical framework developed earlier.
\begin{enumerate}[label=(\roman*),leftmargin=2.2em]
  \item \textbf{A localized decision statistic}: the policy statistic $\pi_t(s)$ that determines the action at a given state $s_t$ is localized to the cell $\sigma(s_t)$ bound to the given state.
  \item \textbf{Amortized transition dynamics}: the relations among cells concentrate on
    the feasible one-step transitions, so the routing \emph{is} a transition model, given by locality in the
    DRC, discovered from reward in the attention core.
  \item \textbf{A decodable plan}: a goal-directed plan is decodable over that hidden-cell graph after the final thinking step, before the next action is taken.
  \item \textbf{Plan refinement with thinking}: the plan is refined with each additional
    thinking step, turning extra computation into a better decision.
\end{enumerate}
These constitute the structural content of decision-time planning: a state-indexed decision (i), read from a learned
model of the dynamics (ii), assembled into a lookahead over reachable futures (iii) that deepens with
computation (iv).

Sections~\ref{sec:emp-localization}--\ref{sec:emp-refine} establish (i)--(iv) on the attention core. Section~\ref{sec:emp-slots} then tests the chain's first link by ablation: an otherwise-matched core that must learn its binding fails to form one, and neither the transition graph nor planning emerges.

\subsection{Action Localization}
\label{sec:emp-localization}
Because the environment is deterministic, selecting an action is selecting a
successor; the statistic that determines the action at $s_t$ is therefore
equivalently a \emph{next-state statistic}, and Assumption~\ref{ass:readout}
predicts that the action statistic is localized at the cell bound to the current state. For the
convolutional core this is established by prior work: concept probes decode,
per square, the direction from which the agent will next enter it and the
direction in which it will next push a box, so
that at the agent's own square the decoded concept is its next move
\citep{bush2025interpreting,taufeeque2024planning}. The attention core, in
which nothing places the decision at the agent's cell a priori, shows the
same signature: a linear probe reading the agent-square cell alone decodes
the model's own next action in $60\%$ of cases, compared with a chance level of $25\%$. The decision
statistic is thus localized in the representation and readable at the current
state's cell, as Section~\ref{sec:localized} requires---though not perfectly.
We suspect the residual gap reflects an undertrained agent rather than a
delocalized readout, and would expect the coupling to tighten alongside
further gains in return with longer training; verifying this lies beyond our
compute budget, and we leave it open.

\subsection{Amortized transition dynamics}
\label{sec:emp-graph}

A convolutional core gets its transition model for free: its kernel passes messages only between grid
neighbours, so information flows only along the grid graph \citep{taufeeque2024planning}. The attention core
is unconstrained, its dense relations can route anywhere, however we find it recovers the transition graph from reward
alone. Its settled attention concentrates on a cell's one-step successors and decays geometrically by a factor $0.61$ with
$k$-hop \emph{graph} reachability, vanishing on cells a wall renders unreachable even one
pixel away (Fig.~\ref{fig:wall}a). We also find the routing to be causal: placing a wall on the agent’s path to the goal, thereby severing an edge it must traverse, shifts the hidden state at the agent's own cell $2.3\times$ more
than an identical off-path wall at matched distance, and flips its greedy move on $19\%$ of route-blocked
boards versus $7.2\%$ off-route (Fig.~\ref{fig:wall}b). Cutting an edge the plan depends on reroutes
information to the decision and rewrites it; cutting an irrelevant one does not. This suggests that the attention-core's routing is an amortized causal transition model of the Sokoban dynamics.

\subsection{A Decodable Plan}
\label{sec:emp-plan}

Choosing the next state from a single cell is one step; planning requires that the choice reflect a
multi-step, goal-directed plan over the graph. In the convolutional core such a plan is decoded directly: the
square-level entry- and push-directions of \citet{bush2025interpreting} assemble into coherent trajectories
from start to goal and transfer to unseen layouts. The attention core carries the same plan as a \emph{field}
over its cells: a linear probe recovers, per cell, a move direction that points towards the goal in $61\%$ of cases (chance $25\%$)
laying the plan out across the whole board rather than only at the agent's square.

\subsection{Plan Refinement with Thinking}
\label{sec:emp-refine}
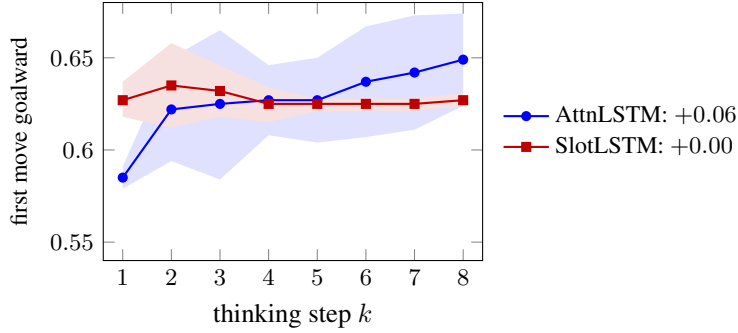
\begin{figure}[t]\centering
\begin{tikzpicture}
\begin{axis}[width=6.6cm,height=5.0cm,xlabel={thinking step $k$},
  ylabel={first move goalward},ylabel style={font=\small},
  xmin=0.6,xmax=8.4,ymin=0.54,ymax=0.68,xtick={1,2,3,4,5,6,7,8},
  legend style={at={(1.03,0.5)},anchor=west,draw=none,font=\small},
  legend cell align=left,tick label style={font=\footnotesize}]
\addplot[draw=none,name path=i4aup,forget plot] coordinates {(1,0.591)(2,0.650)(3,0.665)(4,0.646)(5,0.650)(6,0.667)(7,0.673)(8,0.674)};
\addplot[draw=none,name path=i4alo,forget plot] coordinates {(1,0.579)(2,0.594)(3,0.584)(4,0.608)(5,0.604)(6,0.607)(7,0.611)(8,0.624)};
\addplot[blue!12,forget plot] fill between[of=i4aup and i4alo];
\addplot[draw=none,name path=i4sup,forget plot] coordinates {(1,0.637)(2,0.658)(3,0.646)(4,0.634)(5,0.628)(6,0.628)(7,0.628)(8,0.631)};
\addplot[draw=none,name path=i4slo,forget plot] coordinates {(1,0.618)(2,0.612)(3,0.618)(4,0.615)(5,0.621)(6,0.621)(7,0.621)(8,0.624)};
\addplot[red!75!black!12,forget plot] fill between[of=i4sup and i4slo];
\addplot[mark=*,blue,thick,mark size=1.5pt]
  coordinates {(1,0.585)(2,0.622)(3,0.625)(4,0.627)(5,0.627)(6,0.637)(7,0.642)(8,0.649)};
\addlegendentry{AttnLSTM: ${+}0.06$}
\addplot[mark=square*,red!75!black,thick,mark size=1.5pt]
  coordinates {(1,0.627)(2,0.635)(3,0.632)(4,0.625)(5,0.625)(6,0.625)(7,0.625)(8,0.627)};
\addlegendentry{SlotLSTM: ${+}0.00$}
\end{axis}\end{tikzpicture}
\caption{\textbf{Thinking improves the decision only in the bound core.} At each
thinking step $k$, the greedy action is read from the intermediate hidden state
through the model's own trained actor head, on the initial observation of each of
$512$ held-out boards, and scored by whether the first move is \emph{goalward}. We restrict to hard levels with agent-to-goal distance of $8$--$12$ hops, where lookahead matters most. Here, the attention core keeps improving through the \emph{final} thinking step (${+}0.06$); the slot control is flat (${+}0.00$).}
\label{fig:inert}
\end{figure}

Finally, if these thinking steps perform planning, they should \emph{revise} the decision, the revisions
should \emph{converge}, and they should \emph{improve} the outcome. In the convolutional core, all three hold: the decoded plan concepts sharpen progressively across the recurrent
iterations \citep{bush2025interpreting}. We find that the attention core refines the same way: over its thinking steps it
revises the chosen action on roughly a quarter of boards, and this revision rate falls to $0$ as the decision
settles, while the policy's margin on the chosen action grows $1.7\times$ and the move stays goal-consistent
($\approx\!60\%$ goalward, Fig.~\ref{fig:refine}). The AttnLSTM collects most of this mean gain in the first thinking step, which is to be expected,
since dense routing can span the board in a single step, but on the hardest boards
(agent-to-goal distance $8$--$12$ hops) the gain instead accrues gradually through
the final step (${+}0.06$, Fig.~\ref{fig:inert}), the signature of deepening lookahead.
\footnote{Throughout, a move is \emph{goalward} if it equals the BFS-greedy first step toward the target square, computed over non-wall tiles (chance $0.25$). This is a navigation proxy that ignores box dynamics, as the correct Sokoban move is often box-directed rather than target-directed, so even the settled policy scores only ${\sim}0.6$. For us, goal-directedness is thus used only as a measure of whether a policy is reactive or adaptive under computation, which is a traditional planning signature.}

\subsection{The Binding is Load-Bearing: a Free-Slot Control}
\label{sec:emp-slots}

Both cores explored until now share one strong architectural prior: the convolutional encoder pins each board
square to its own cell, so the binding $\sigma$ of Definition~\ref{def:binding} is
supplied by construction and only the relations must be learned. This leaves open which ingredient the emergence actually rests on: just the relational message passing,
or the message passing \emph{over that given binding}. To separate them, we train
a third, otherwise-matched core in which the binding itself must be discovered. In
the \emph{slot} core (SlotLSTM), the $100$ cells are free slots with no
positional identity: at every thinking step each slot first \emph{binds} by a
slot-attention competition over the encoder's board tokens (a softmax across
slots, so the slots compete to explain each square), then \emph{routes} by the same dense slot-to-slot entmax attention as the AttnLSTM,
and integrates both messages through the same LSTM gate (App.~\ref{app:arch}). Position remains available as content, the board tokens carry a learned positional embedding, so what
is removed is exactly the fixed cell-to-square correspondence, nothing else.

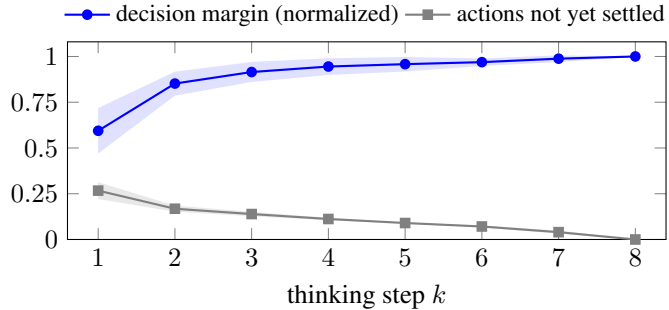
\begin{figure}[t]\centering
\begin{tikzpicture}
\begin{axis}[width=9.5cm,height=4.2cm,xlabel={thinking step $k$},
  xmin=0.6,xmax=8.4,ymin=0,ymax=1.08,xtick={1,2,3,4,5,6,7,8},
  ytick={0,0.25,0.5,0.75,1},
  legend style={at={(0.5,1.03)},anchor=south,legend columns=-1,draw=none,font=\small}]
\addplot[draw=none,name path=r5mup,forget plot] coordinates {(1,0.718)(2,0.917)(3,0.970)(4,0.990)(5,0.997)(6,0.989)(7,1.006)(8,1.000)};
\addplot[draw=none,name path=r5mlo,forget plot] coordinates {(1,0.470)(2,0.786)(3,0.861)(4,0.899)(5,0.918)(6,0.948)(7,0.970)(8,1.000)};
\addplot[blue!12,forget plot] fill between[of=r5mup and r5mlo];
\addplot[draw=none,name path=r5sup,forget plot] coordinates {(1,0.313)(2,0.185)(3,0.151)(4,0.118)(5,0.093)(6,0.076)(7,0.049)(8,0.000)};
\addplot[draw=none,name path=r5slo,forget plot] coordinates {(1,0.221)(2,0.152)(3,0.127)(4,0.105)(5,0.087)(6,0.066)(7,0.032)(8,0.000)};
\addplot[gray!18,forget plot] fill between[of=r5sup and r5slo];
\addplot[mark=*,blue,thick,mark size=1.6pt]
  coordinates {(1,0.594)(2,0.852)(3,0.915)(4,0.945)(5,0.958)(6,0.969)(7,0.988)(8,1.000)};
\addlegendentry{decision margin (normalized)}
\addplot[mark=square*,gray,thick,mark size=1.6pt]
  coordinates {(1,0.267)(2,0.168)(3,0.139)(4,0.112)(5,0.090)(6,0.071)(7,0.040)(8,0.000)};
\addlegendentry{actions not yet settled}
\end{axis}\end{tikzpicture}
\caption{\textbf{Each thinking step refines the decision.} At every thinking step $k$ of the AttnLSTM we
read the policy from the intermediate state $h_t^{k}$ through the trained actor
head. The margin on the chosen action, the top-1 minus top-2 action logit,
is shown normalized to its converged
value (blue) and rises monotonically to $1$, while the fraction of boards whose
greedy action has not yet reached its settled value (grey) falls from $27\%$
to $0$: the decision sharpens and converges by step $8$.}
\label{fig:refine}
\end{figure}

\textbf{No state-graph binding emerges.} Definition~\ref{def:binding} asks for a
map from states to cells under which each state's statistic is carried by its
own, recoverable cell, and, for the weight-tied routing to use it, a map that
holds across boards. We find that the slot core does not discover this. Its binding reads are diffuse rather than one-to-one: a slot places only $0.13$ of its read mass on its top
square, $3$--$4$ slots share each bound square, and only $43\%$ of navigable
squares are any slot's primary read. Furthermore, the assignment is not stable across Sokoban levels and even the most agent-specialized
slot reads the agent on only $29\%$ of boards. However, the slots are not empty: a square's tile still decodes from its winning slot at
$0.66$ vs.\ $0.50$ from a random one; what fails to form is the unique,
board-independent correspondence Def.~\ref{def:binding} requires.

\textbf{Without binding, neither a transition graph nor planning emerges.}
While the anchored attention core's routing decays geometrically with graph
distance between the squares its cells hold ($\rho\approx0.61$ per hop,
Fig.~\ref{fig:wall}a), the slot core's is nearly distance-blind
($\rho\approx0.94$): $67\%$ of its mass connects squares more than one
transition apart, it carries no asymmetry toward the goal, and a perturbation
at \emph{any} graph distance $1$--$9$ reaches the agent's slot within the first
thinking step (arrival slope $0.00$ steps per hop). The broadcast still delivers information: a route-blocking wall shifts the decoded
value at the agent's slot $2.9\times$ more than a matched off-route wall, a ratio at least as large as the attention core's. But this shift arrives instantaneously, and extra
computation yields no better decision, the opposite of what planning predicts.
Read through the actor head at each step, the first move's goalward fraction is flat
(${+}0.004$ from step $1$ to $8$, vs.\ ${+}0.06$ for the AttnLSTM), the move is
revised on $9\%$ of boards (vs.\ $27\%$, Fig.~\ref{fig:refine}), the
route-blocking wall flips the move on only $11\%$ of route-blocked boards (vs.\
$19\%$, Fig.~\ref{fig:wall}b), and eight extra recurrent steps on the initial
observation (the pondering protocol of \citealp{guez2019investigation}) leave the slot core's held-out solve rate essentially unchanged ($0.067$ with $0$ extra steps, $0.063$ with $8$); under the
same pondering protocol the DRC gains $0.754\to0.805$ (Fig.~\ref{fig:solve}a),
while the AttnLSTM under its inner-tick sweep gains $0.31\to0.39$
(Fig.~\ref{fig:solve}b).
We further investigated whether the difference in variable-depth and fixed-depth computation could be at fault here. However, we found that an AttnLSTM trained at the same
$K{=}4$ gains ${+}0.080$ in goal-directedness gradually over steps $1$--$5$, whereas the slot core is
flat within \emph{every} distance band (${\pm}0.03$, sign-incoherent). We further find that a $50$-slot variant replicates every finding: routing barely decays over successors ($\rho\approx0.97$), and the  solve rate increases marginally at best ($0.051\to0.059$).

\section{Discussion \& Future Work}
\label{sec:discussion}

Our results show that a neural network with a graph-like backbone can be trained
to learn decision-time planning from reward maximization alone, without an
explicit model or a prescribed search procedure. Because the convolutional and attention-based relational cores we
study are common among modern deep-learning architectures, this raises the possibility that emergent
planning is a fairly general phenomenon in reinforcement learning. Several works have investigated whether planning mechanisms can emerge in transformers, and some have found indications that they do, such as search- and lookahead-like computation across
Othello and chess transformers \citep{li2023othello,jenner2024chess} and emergent planning in language
models \citep{lindsey2025biology,dong2025emergent}.

We caution against over-generalizing: in both planning cores, each environment state is pinned to a single cell by a convolutional encoder, and only the hidden-cell relations are learned. The slot control (\S\ref{sec:emp-slots}) shows that this strong inductive bias is not incidental but load-bearing: when the binding must itself be discovered, our agent fails to form one, only a diffuse, per-board assignment emerges. From this, no transition structure settles into the networks' relations, and planning does not emerge. Most large-scale models lack the one-state-per-cell structure. For example, a transformer's residual stream mixes many features per position, and positions need not correspond to environment states at all. On our account, the operative question for such models is whether a stable-enough binding is realized elsewhere in their representations, and whether planning emerges exactly where it is: emergent planning may be common in architectures that supply this binding cheaply and rarer elsewhere. Whether a different learned-binding mechanism, longer training, or greater scale can stabilize $\sigma$ where slot competition did not, we regard as open and leave to further research.

Our empirical claims rest on three independently trained AttnLSTMs and three
independently trained slot controls (the original run plus two additional random
seeds each), together with the single pretrained DRC
checkpoint~\citep{bush2025interpreting}, evaluated on one test set. Across the three
AttnLSTM seeds the mechanism is stable rather than a single-seed artifact: the
routing recovers the transition graph in every run (graph-distance decay
$\rho = 0.61$, and blocked-by-wall influence far below the open baseline in
all three), the decision statistic stays localized to the agent's cell
($60\%$), and the decision measurably refines over thinking steps in each run
(first-step revision rate $27\%$, falling to zero as it settles). The negative
result is equally stable: across the three slot controls the binding stays diffuse
($0.13$ read mass on the top square; the most agent-specialized slot reads
the agent on $29\%$ of boards) and the routing stays a near-uniform broadcast
($\rho = 0.94$, $67\%$ of mass beyond one hop, arrival slope $0.00$),
with no plan refinement. Point estimates in the main text are 3-seed means; the
per-seed values and standard deviations for every reported quantity are given in
Table~\ref{tab:seeds} (App.~\ref{app:seeds}), and the figures show the corresponding
mean$\pm$sd bands. The binding nonetheless remains architectural in the
AttnLSTM and ConvLSTM cores---the convolutional encoder pins each square to its own
cell---so it is not itself an emergent quantity; and a systematic sweep over
initialization scale, training length, and alternative learned-binding mechanisms
that might stabilize $\sigma$ where slot competition did not remains future work.

We close by noting that many accounts of how the human brain plans rely on a similar relational-graph mechanism. In cognitive science, the hippocampus is often described as a relational graph \citep{tolman1948cognitive,
stachenfeld2017hippocampus,whittington2020tolman} that binds abstract elements by
their relations \citep{konkel2009relational,eichenbaum2017integration}. If a biological system already supplies a relational state graph, our account suggests reward-driven learning over it could yield planning by the same route. However, we offer this only as a
hypothesis raised by our framework, not a claim we establish.

\section{Acknowledgements}
The author is supported by an Excellence Scholarship from the State of Lower Austria. The author would like to express his gratitude to Jannik Schilling for his feedback and suggestions on initial drafts.

\bibliographystyle{plainnat}
\bibliography{references}

\newpage
\appendix
\onecolumn

\section{Appendix: Architecture and Training Details}
\label{app:arch}

\paragraph{Shared relational recurrence.}
A convolutional encoder of two $4\times4$ same-padded conv layers, no nonlinearity,
$C{=}32$ channels; \citealp{guez2019investigation} maps the observation to a
feature map $e\in\mathbb{R}^{H\times W\times C}$. Here $H{=}W{=}10$ is read as
$S{=}HW{=}100$ cells indexed by board position with the playable region being the inner
$8\times8$. A core of $D$ stacked, weight-tied cells is applied for $K$
``thinking'' ticks within one environment step $t$,
and the settled state is carried across steps, $h_t^{0}\equiv h_{t-1}^{K}$. Each
cell layer $d$ holds an LSTM state $(c_d,h_d)$ and, at repeat $k$, updates it from
its own previous state $h_d^{k-1}$, the encoder features $e$, and the layer below
$h_{d-1}^{k}$, via a relation over the $S$ cells. The two instantiations below
differ only in that relation.

\paragraph{Convolutional core: ConvLSTM \citep{guez2019investigation}.}
The relation is a fixed local kernel, realised by a convolutional LSTM
\citep{shi2015convlstm}. The layer-$d$ input at repeat $k$ concatenates the
encoder features, the hidden state below, a pool-and-inject summary, and a
boundary-indicator channel,
\begin{align}
x_d^{k} &\;=\; \bigl[\,e\,;\; h_{d-1}^{k}\,;\; \mathrm{p}(h_d^{k-1})\,;\; \mathbf 1_{\partial}\,\bigr],\\
\mathrm{p}(h) &\;=\; \mathrm{broadcast}\!\big(W_p[\,\mathrm{maxpool}_{HW}h\,;\,\mathrm{meanpool}_{HW}h\,]\big),
\end{align}
where $\mathbf 1_{\partial}$ marks the grid boundary (ones on the border, zeros
inside; not zero-padded) to keep convolution edge effects from corrupting the
recurrence. A $3\times3$ ConvLSTM then updates the cell, with $*$ a $2$-D
convolution and $\odot$ the Hadamard product:
\begin{align}
i &= \sigma\!\big(W_i * x_d^{k} + U_i * h_d^{k-1}\big),\\
f &= \sigma\!\big(W_f * x_d^{k} + U_f * h_d^{k-1} + b_f\big),\\
o &= \tanh\!\big(W_o * x_d^{k} + U_o * h_d^{k-1}\big),\\
g &= \tanh\!\big(W_g * x_d^{k} + U_g * h_d^{k-1}\big),\\
c_d^{k} &= f\odot c_d^{k-1} + i\odot g,\\
h_d^{k} &= o\odot\tanh\!\big(c_d^{k}\big).
\end{align}
The $\tanh$ output gate follows \citet{jozefowicz2015empirical} and the forget
bias $b_f$ is initialised to $1$. Because messages are gathered by the $3\times3$
kernel, each cell $s$ draws only from its eight spatial neighbours; the
transition graph is fixed by the architecture, while pool-and-inject adds a
coarse global summary (a whole-board max/mean), not a learned pairwise relation.
We use $D{=}3$, $K{=}3$ (the DRC$(3,3)$ of \citealp{guez2019investigation}) and
the pretrained checkpoint of \citet{bush2025interpreting}.

\paragraph{Attention core (AttnLSTM).}
The relation is implemented as learned, content-based attention over all $S$ cells (no locality
mask). With $\tilde h_s=\mathrm{RMSNorm}\!\big(h_s + W_{\text{in}}[\,e_s;\,h_{s,\text{below}}\,]\big)$
projected to $n_h$ heads of width $d_h=C/n_h$,
\begin{align}
q_s,\;k_s,\;v_s &= W_q\tilde h_s,\; W_k\tilde h_s,\; W_v\tilde h_s,\\
L_{s,r} &= \tfrac{1}{\sqrt{d_h}}\,q_s^{\!\top}k_r + b_{\,s-r},\\
w_{s,\cdot} &= \operatorname{entmax}_{1.5}\!\big(L_{s,\cdot}\big),\\
a_s &= W_o \textstyle\sum_{r} w_{s,r}\,v_r,
\end{align}
where $b_{\,s-r}$ is an offset-tied relative-position bias (zero-initialised).
Thus, $w_{s,\cdot}$ is a learned, $1.5$-entmax-sparse distribution over the cells
from which $s$ draws; its support defines the recovered graph $\mathcal N$. The
message $a_s$ drives the same LSTM gates as above with $x_d^{k}=[\,e_s;\,a_s\,]$, with $1.5$-entmax being an architectural choice to incentivize sparse routing in the attention. Our experiments use the default hyperparameters of $D{=}3$, $n_h{=}4$, $C{=}32$, with thinking depth $K$ ranging up to $6$.

\paragraph{Slot core (SlotLSTM).}
The SlotLSTM replaces the $S$ position-indexed cells with $N{=}100$ free slots
with no spatial identity. The carry between recurrent steps is $(c,h)\in\mathbb R^{N\times C}$, reset to zero
at episode boundaries; a learned per-slot identity $\mu_i$---the only thing
distinguishing slot $i$ from slot $j$---enters additively at every step,
$\tilde h=\mathrm{RMSNorm}(h+\mu)$. Each thinking step computes two messages. The
\emph{binding} read queries the $S$ encoder tokens (which carry a learned
positional embedding, so position is available as content): with
$q_i=W^b_q\tilde h_i$ and $k_s,v_s=W^b_k e_s,\,W^b_v e_s$,
\begin{align}
A_{is} \;=\; \frac{\exp\!\big(q_i^{\!\top}k_s/\sqrt{d_h}\big)}{\sum_{i'}\exp\!\big(q_{i'}^{\!\top}k_s/\sqrt{d_h}\big)},
\qquad
m^{\text{bind}}_i \;=\; W^b_o \textstyle\sum_s \bar A_{is}\, v_s,
\end{align}
where the softmax runs over \emph{slots}. This can be seen as the slot-attention competition: slots
compete to explain each square, which breaks the permutation symmetry and ties
slots to states, with $\bar A$ renormalizing each slot's row over the tokens. The
\emph{routing} message is the same dense, multi-head $1.5$-entmax attention as
the attention cell, applied slot-to-slot ($q,k,v$ from $\tilde h$; no mask and no
positional term).
Both messages drive the same LSTM gates with
$x_d^{k}=[\,m^{\text{bind}};\,m^{\text{route}}\,]$, and the same flatten-MLP
readout is applied to the slot field. We use $N{=}100$ (matching the $100$ board
squares), $D{=}3$, $n_h{=}4$, $C{=}32$, at a fixed thinking depth $K{=}4$,
trained with the identical IMPALA recipe for $3\times10^{8}$ steps (a $N{=}50$
variant behaves the same throughout \S\ref{sec:emp-slots}). The undertaken measurements are: \emph{binding}, by decoding a square's tile from the slot whose
binding read $\bar A$ peaks on that square (against a random slot), and by the
across-board consistency of that assignment; \emph{routing}, by binning the
slot-to-slot weights by the graph distance between the squares the two slots
currently bind; \emph{reach}, by the arrival step of a perturbation's effect at
the agent's slot as a function of graph distance; and \emph{decision refinement},
by the per-step actor readout of Figs.~\ref{fig:refine} and~\ref{fig:inert} and
by the solve rate with $0$ vs.\ $8$ extra recurrent steps on the initial
observation.

\paragraph{Readout.}
After $K$ repeats, the top layer's settled state, with the encoder skip, is
flattened and passed through a one-hidden-layer MLP ($256$ units) shared by actor
and critic,
\begin{align}
u &= \mathrm{relu}\!\big(W_d\,\mathrm{vec}(h^{K}_{\text{top}}+e)+b_d\big),\\
\pi(\cdot\mid s_t) &= \mathrm{softmax}(W_a u),\\
V &= W_c u.
\end{align}

\paragraph{Training.}
We train the attention core with IMPALA, an actor–critic method with V-trace, on Boxoban unfiltered-train for $3\times10^8$ environment steps, using $\gamma=0.97$, gradient clipping, and entropy regularisation \cite{espeholt2018impala}. The attention core is trained with \emph{variable thinking
depth}: each rollout samples $K\sim\mathcal U\{1,\dots,6\}$ and the learner
replays the same depth, so the agent must act at every budget (a fixed-depth
$K{=}4$ variant is used where noted). No model-prediction or reconstruction
auxiliary loss is used; the transition graph is shaped by reward alone.

We trained the AttnLSTM for 300 million environment steps, achieving a $91.9\%$ solve rate on the training set. The ConvLSTM was pretrained by Bush et al.~\citep{bush2025interpreting} for 2 billion environment steps, achieving a $98.4\%$ solve rate on the training set. The slot control was trained with the same recipe and budget at fixed $K{=}4$, reaching a $65$--$68\%$ training solve rate.

\paragraph{Faithful recompute.}
The model's forward pass triggers a tracer error on the offset-tied relative-bias
gather under \texttt{nn.scan} (JIT/GPU). All interpretability probes therefore
recompute the $D{=}3$ attention stack in plain JAX from the trained parameters,
and apply the head matrix multiplications for logits and value; this reproduces the hidden state to within $1.8\times10^{-7}$ and is used for all results in Section \ref{sec:emp}.

\section{Per-Seed Values and Run-to-Run Variance}
\label{app:seeds}

Every quantitative claim in the main text is reported as the mean over three
independently trained models: the original run (seed $4242$) and two additional
random seeds ($1$, $2$). Table~\ref{tab:seeds} lists the per-seed value and the
mean\,$\pm$\,standard deviation for each reported quantity; the corresponding
bands are drawn on Figs.~\ref{fig:solve},~\ref{fig:wall},~\ref{fig:inert}
and~\ref{fig:refine}. Each probe uses the same protocol across seeds (and the
original seed reproduces the previously reported single-seed value). The pretrained
DRC$(3,3)$ is a single external checkpoint and is not seed-varied.

\begin{table}[h]
\centering
\small
\begin{tabular}{@{}llcccc@{}}
\toprule
Quantity & source & seed 4242 & seed 1 & seed 2 & mean $\pm$ sd \\
\midrule
\multicolumn{6}{@{}l}{\textit{AttnLSTM --- emergent planning}}\\
Action decode @ agent cell (chance $0.25$) & \S\ref{sec:emp-localization} & 0.615 & 0.577 & 0.615 & $0.60 \pm 0.02$ \\
Routing decay $\rho$ (graph distance)       & \S\ref{sec:emp-graph}, Fig.~\ref{fig:wall}a & 0.659 & 0.642 & 0.529 & $0.61 \pm 0.06$ \\
One-hop attention mass                       & Fig.~\ref{fig:wall}a & 0.206 & 0.244 & 0.256 & $0.24 \pm 0.02$ \\
Wall$\to$agent shift ratio                   & \S\ref{sec:emp-graph}, Fig.~\ref{fig:wall}b & 2.23 & 2.39 & 2.41 & $2.34 \pm 0.08$ \\
On-route flip rate                           & Fig.~\ref{fig:wall}b & 0.200 & 0.183 & 0.200 & $0.19 \pm 0.01$ \\
Off-route flip rate                          & Fig.~\ref{fig:wall}b & 0.067 & 0.067 & 0.083 & $0.072 \pm 0.008$ \\
Hard-band goalward gain over ticks           & \S\ref{sec:emp-refine}, Fig.~\ref{fig:inert} & $+0.081$ & $+0.022$ & $+0.089$ & $+0.06 \pm 0.03$ \\
First-step revision rate                     & \S\ref{sec:emp-refine}, Fig.~\ref{fig:refine} & 0.346 & 0.244 & 0.234 & $0.27 \pm 0.05$ \\
Thinking-curve solve rate (peak)             & Fig.~\ref{fig:solve}b & 0.421 & 0.394 & 0.368 & $0.39 \pm 0.02$ \\
Training-set solve rate                       & App.~\ref{app:arch} & 0.923 & 0.917 & 0.916 & $0.919 \pm 0.003$ \\
\midrule
\multicolumn{6}{@{}l}{\textit{SlotLSTM --- free-slot control}}\\
Routing decay $\rho$ (near-uniform)          & \S\ref{sec:emp-slots} & 0.946 & 0.959 & 0.912 & $0.94 \pm 0.02$ \\
Routing mass beyond one hop                  & \S\ref{sec:emp-slots} & 0.721 & 0.649 & 0.651 & $0.67 \pm 0.03$ \\
Winning-slot tile decode (chance $0.50$)     & \S\ref{sec:emp-slots} & 0.665 & 0.675 & 0.624 & $0.66 \pm 0.02$ \\
Read mass on top square                      & \S\ref{sec:emp-slots} & 0.112 & 0.149 & 0.118 & $0.13 \pm 0.02$ \\
Agent-slot stability across boards           & \S\ref{sec:emp-slots} & 0.32 & 0.28 & 0.27 & $0.29 \pm 0.02$ \\
Wall$\to$slot value-shift ratio              & \S\ref{sec:emp-slots} & 2.43 & 3.50 & 2.65 & $2.9 \pm 0.5$ \\
Flip rate                                    & \S\ref{sec:emp-slots} & 0.070 & 0.083 & 0.167 & $0.11 \pm 0.04$ \\
Goalward gain over ticks                      & \S\ref{sec:emp-slots}, Fig.~\ref{fig:inert} & $+0.002$ & $+0.010$ & $+0.000$ & $+0.004 \pm 0.004$ \\
First-step revision rate                     & \S\ref{sec:emp-slots}, Fig.~\ref{fig:refine} & 0.102 & 0.061 & 0.119 & $0.09 \pm 0.02$ \\
Pondering solve rate ($0$ steps)             & \S\ref{sec:emp-slots} & 0.073 & 0.051 & 0.076 & $0.067 \pm 0.011$ \\
Pondering solve rate ($8$ steps)             & \S\ref{sec:emp-slots} & 0.070 & 0.051 & 0.069 & $0.063 \pm 0.009$ \\
Training-set solve rate                       & App.~\ref{app:arch} & 0.681 & 0.648 & 0.672 & $0.667 \pm 0.014$ \\
\bottomrule
\end{tabular}
\caption{\textbf{Per-seed values and run-to-run variance} over three independently
trained models. Main-text numbers are the mean column; the figures show mean$\pm$sd
bands. The AttnLSTM planning signatures and the SlotLSTM control's failure both hold
across all three seeds.}
\label{tab:seeds}
\end{table}

\end{document}